\documentclass{ai-tutor-sigconf}

\AtBeginDocument{%
  }

\usepackage[inkscapeformat=png]{svg}
\usepackage{CJKutf8}

\usepackage{enumitem}
\usepackage{tikz}
\usetikzlibrary{shapes.geometric, arrows.meta, positioning, calc}

\begin{document}

\title{Multi-sourced AI Tutor for Non-Native English Speaker Students}

\author{%
\begin{tabular}{ccc}
\begin{minipage}[t]{0.30\textwidth}\centering
Yiyin Shen\\University of Wisconsin--Madison\\Madison, WI, USA\\\url{yshen82@wisc.edu}
\end{minipage} &
\begin{minipage}[t]{0.30\textwidth}\centering
Xinyi Ai\\Independent Researcher\\Mountain View, CA, USA\\\url{cathyaxy@gmail.com}
\end{minipage} &
\begin{minipage}[t]{0.30\textwidth}\centering
Weibing Wang\\Massachusetts Institute of Technology\\Cambridge, MA, USA\\\url{weibingw@mit.edu}
\end{minipage}
\end{tabular}
\par\vspace{1em}
\begin{tabular}{cc}
\begin{minipage}[t]{0.30\textwidth}\centering
Rogers Jeffrey Leo John\\Independent Researcher\\Madison, WI, USA\\\url{rogersjeffreyl@gmail.com}
\end{minipage} &
\begin{minipage}[t]{0.30\textwidth}\centering
Meenakshi Syamkumar\\University of Wisconsin--Madison\\Madison, WI, USA\\\url{ms@cs.wisc.edu}
\end{minipage}
\end{tabular}}

\begin{abstract}
Large language models (LLMs) are increasingly being adopted as teaching tools in computing education, particularly for developing AI tutors capable of answering student questions. With the rise in enrollment of international students in computing programs, researchers have begun to explore the difficulties that non-native English speaker (NNES) students face in computing courses due to language barriers. To aid such students, we developed a general-purpose AI tutor system that can process student questions in foreign languages (currently focused on Mandarin) and provide bilingual responses. Our system also incorporates the functionality of generating a glossary of technical terms and their explanations to enhance comprehension. Our approach utilizes a specialized Retrieval-Augmented Generation (RAG) pipeline with a novel question extraction mechanism, enabling us to effectively utilize a diverse array of course materials for answering student queries. We evaluated our AI tutor using 91 Piazza posts, demonstrating that it could answer 83.52\% of the posts correctly and in a helpful manner. The tutor also provided high-quality translations for 76.92\% of the questions and generated technical term glossaries for 64.47\% of them. Our findings underscore the potential of multi-source RAG pipeline in enhancing AI tutor systems, offering support for non-native speakers that are challenging to provide with traditional supporting staff of a computing course.
\end{abstract}

\keywords{large language model, non-native English speaker students, computing education}
\maketitle

\vspace{-0.4em}

\section{Introduction}
Large Language Models (LLMs) have profoundly transformed computing education, showcasing remarkable capabilities in solving computing assignments \cite{10.1145/3626252.3630874, 10.1145/3633287}. LLMs are also being used for facilitating instructional activities, such as generating assignment questions \cite{sarsa2022automatic}, providing automated grading \cite{xie2024gradelikehumanrethinking}, and notably, developing AI tutor systems that address student queries. Furthermore, as classrooms become increasingly multicultural, the need for inclusivity in educational materials and technologies grows. Non-native English speaker (NNES) students, in particular, face language barriers and a lack of confidence, which often causes hesitation to seek help from instructors or TAs. Our AI tutor system seeks to assist NNES students' learning by accepting questions in the students' native language (specifically Mandarin), delivering responses bilingually, and generating a technical term glossary to ensure equitable access to educational materials.

Most AI tutors \cite{10.1145/3626252.3630789, 10.1145/3626252.3630938, hicke2023aita} utilize Retrieval-Augmented Generation (RAG) \cite{NEURIPS2020_6b493230}, which enhances question-answering by dynamically retrieving and synthesizing relevant information from a data repository, ensuring accuracy and contextual relevance. Our RAG-based GPT-4 tutor differs from systems that rely on a single source, such as lecture captions or textbooks. Instead, it integrates a comprehensive range of course materials, including lecture notes, project descriptions, quizzes, past exams, and the course website, enabling holistic data retrieval.

Two specialized mechanisms have been developed to enhance our AI tutor. Firstly, our system incorporates a translation stage that converts Mandarin questions into English before applying RAG, enabling full leverage of RAG's capabilities. Secondly, our system employs a question extraction mechanism that addresses instances where a student's query references only the question number instead of providing the question details, by precisely identifying and retrieving details of the referred question.

We evaluated our system using 91 Piazza posts, demonstrating that it can accurately and helpfully answer student questions and generate technical term glossary in Mandarin. Our work focuses on answering the following research questions: 
\begin{itemize}[leftmargin=1.5em]
    \item \textbf{RQ1}. What is the quality of the translation stage and to what extent does it improve RAG's performance in processing non-English questions?
    \item \textbf{RQ2}. What is the accuracy of the question extraction stage and to what extent does it enhance the system's ability to manage diverse course materials and student questions?
    \item \textbf{RQ3}. What is the relevance and usefulness of RAG's results concerning the questions?
    \item \textbf{RQ4}. What is the quality of the AI tutor's responses in terms of correctness, helpfulness, solution revelation, translation, and technical term glossary generation?
\end{itemize}


\vspace{-0.5em}

\section{Related Work}
\subsection{AI Tutors}
Since the debut of Large Language Models (LLMs), researchers have consistently explored their potential applications in computing education. Beyond aiding student learning by providing thorough code explanations \cite{macneil2023experiences}, step-wise worked examples \cite{10.1145/3636243.3636252}, and prompt problems \cite{10.1145/3626252.3630909}, as well as facilitating various instructional activities, LLMs have proven effective as the basis for educational question-answering (QA) systems, or in other words, AI tutors. 

Hicke et al. \cite{hicke2023aita} investigated the effectiveness of various optimization techniques in LLMs for answering questions collected from a CS1 course Piazza. They enhanced Llama-2 using Retrieval-Augmented Generation (RAG) on textbooks and lab manuals, supervised finetuning with selected Piazza QA pairs, and reinforcement learning with human feedback; and applied RAG to GPT-4. They show that these techniques resulted in substantial improvements: Llama-2's answer accuracy (factual correctness \& groundedness) increased from 0.30 to 0.58, and GPT-4's from 0.40 to 0.73, while usefulness (practical values to human instructors) increased from 0.18 to 0.50 for Llama-2 and from 0.28 to 0.65 for GPT-4.

Liu and M'hiri \cite{10.1145/3626252.3630789} developed a virtual teaching assistant (TA) using GPT-3.5 and RAG on assignment documents. 
The virtual TA first performs embedding searches to compare students' questions with assignment documents, preventing the revelation of answers to the current assignments to students who directly ask for them. Subsequently, the TA generates responses tailored to each question category and conducts quality checks to ensure response accuracy and avoid solution leak.

Furthermore, 
Liu et al. \cite{10.1145/3626252.3630938} and Kazemitabaar et al. \cite{10.1145/3613904.3642773} developed their own QA chatbots and deployed them in real course settings. The two chatbots have distinct features: the prior utilized GPT-4 with RAG on lecture captioning and structured prompts to generate responses, while the latter used the few-shot learning framework to prevent code revelation. Both chatbots maintained a high correctness rate in resolving student inquires throughout the semester. Moreover, students expressed satisfaction with these tools in the feedback survey, stating they appreciated having such a "24/7 personal tutor" and the opportunity to "work alongside AI."

Most of the AI tutor systems above use RAG based on a single source format, such as textbooks, assignment documents, or lecture captions. Our AI tutor aims to incorporate all documents in a course, including lecture materials and readings, lab and project description documents, quizzes and past exams, as well as the course website, by developing appropriate data pipelines and prompt techniques for each type of course material.


\subsection{Non-Native English Speakers in Computing Education}

The enrollment of international students in computing programs has increased by 14\% to 298,523 during the 2022-23 academic year. Notably, 23\% of them pursued mathematics and computer science, and 19\% studied engineering \cite{opendoors2023}. These students often face language barriers and confidence issues, which negatively impact their learning experience in computing courses. In response, researchers are exploring tailored supportive strategies for this demographic.

Polishchuk \cite{polishchuk2017communication} surveyed 124 first-year non-native English speaker (NNES) engineering and agriculture students,
identifying several key communication barriers, including vocabulary and grammar difficulties 
and challenges with accents and slang. Notably, 85.5\% of students reported translating utterances to their native language, which hindered their fluency, while 61.25\% were afraid of making mistakes, and 47.5\% felt nervous speaking in front of the class. Polishchuk recommended enhancing collaborative work and communication practice outside the classroom.

Guo \cite{inproceedings} noted how confidence issues affect NNES's communication with instructors and TAs.
Via a survey of 840 respondents from 86 countries on a programming education website, this study found students reluctant to participate in discussions or ask questions due to fear of errors and being misunderstood. They preferred instructional materials in simplified English, supplemented with visuals 
and inline dictionaries. Guo suggested peer mentoring and language support workshops to improve participation and confidence.

Soosai Raj et al. \cite{raj2017students, raj2019impact} conducted a four-week experiment with second-year engineering students in India, comparing those taught only using English to those receiving bilingual instruction in English and Tamil. The study found that bilingual teaching reduced cognitive load, improved comprehension of programming concepts, and boosted student engagement and confidence. Lau and Yuen \cite{lau2011impact} noted that K-12 students in Hong Kong understood programming better with bilingual instruction, highlighting the value of native language incorporation for NNES students.

To address the challenges faced by NNES students, we aim to develop an AI tutor that accepts questions in the student's native language and provides bilingual answers to facilitate understanding of complex computing concepts. This approach will also help build confidence among students who may hesitate to approach their instructors and TAs. Since our authors are mainly native Mandarin speakers, we implemented our prototypical system for Mandarin.

\section{Method}
\subsection{Students' Questions}

We collected 620 questions from a semester's Piazza posts for an intermediate-level Data Science (DS) programming course of 400 students\footnote{IRB details will be included upon publication.}.
We excluded 238 posts that included images for two reasons: (1) they are difficult to transcribe, either because they are entire desktop screenshots with only a small portion containing relevant information or they involve complex plots and graphs; (2) the GPT-4 API vision service was prohibitively expensive prior to the release of GPT-4o. The analysis focused on the remaining 382 text-based posts, which we cleaned by removing duplicates, identifiable information, unnecessary HTML formatting, and enclosing code snippets and error messages in backticks.

For Piazza posts, students must tag their posts with instructor-defined folders: exams (exam 1, exam 2, final), quizzes, programming projects (P1, P2, ..., P7), labs, logistics, and others. One author used these student-tagged folders and the post content to sort the posts into four categories: 90 logistics, 54 conceptual, 41 environment setup, and 197 programming posts. Logistic questions dealt with syllabus specifics and demands for clarifications on acceptable coding practices and plotting requirements. Conceptual questions typically pertained to quiz and exam content. Environment setup questions covered the setup and maintenance of virtual machines and JupyterLab, package installations, and git-related issues. Programming questions primarily revolved around project development challenges, addressing issues such as implementation of specific functionalities, debugging, and optimization for code efficiency.

We then generated the Mandarin dataset by translating the Piazza posts into Mandarin using the Google Translate API\footnote{The translation component is modular and can be replaced with other APIs.}. To ensure precision, all translations were manually reviewed and edited by a native Mandarin-speaking author. Sections previously marked with backticks were excluded from the translation.


\tikzstyle{startstop} = [rectangle, rounded corners, minimum width=3cm, minimum height=0.8cm,text centered, draw=black, fill=red!30]
\tikzstyle{process} = [rectangle, minimum width=3cm, minimum height=0.8cm, text centered, draw=black, fill=orange!30]
\tikzstyle{decision} = [diamond, minimum width=4cm, minimum height=0.8cm, text centered, draw=black, fill=green!30, aspect=3]
\tikzstyle{arrow} = [thick,->,>=stealth]

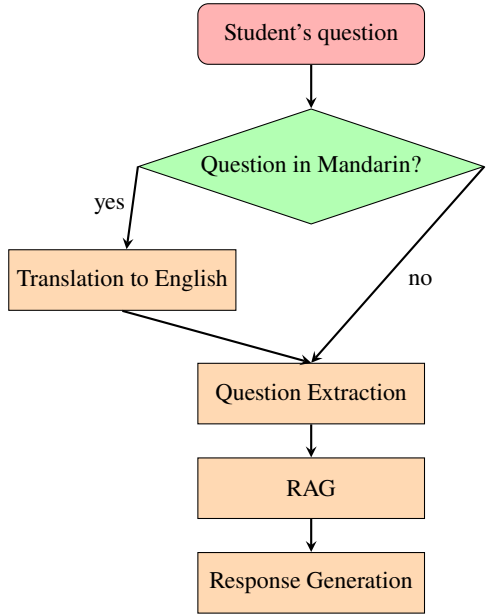
\begin{figure}
    \centering
    \begin{tikzpicture}[node distance=1cm]
        \node (start) [startstop] at (0, 0) {Student's question};
        \node (dec1) [decision] at (0, -1.75) {Question in Mandarin?};
        \node (pro1) [process] at (-2.5, -3.25) {Translation to English};
        \node (pro2) [process] at (0, -4.75) {Question Extraction};
        \node (pro3) [process] at (0, -6) {RAG};
        \node (out1) [process] at (0, -7.25) {Response Generation};
        
        \draw [arrow] (start) -- (dec1);
        \draw [arrow] (pro1.south) -- (pro2.north);
        \draw [arrow] (pro2) -- (pro3);
        \draw [arrow] (pro3) -- (out1);
        \coordinate (leftbottom) at ($(dec1.south)!0.5!(dec1.west)$); 
        \draw [arrow] (dec1.west) -- node[anchor=south,xshift=-0.3cm, yshift=-0.2cm] {yes} (pro1);
        \coordinate (rightbottom) at ($(dec1.south)!0.5!(dec1.east)$); 
        \draw [arrow] (dec1.east) -- node[anchor=north, xshift=0.3cm] {no} (pro2.north);
    \end{tikzpicture}
    \caption{Flowchart of the RAG-based pipeline}
    \label{fig:flowchat}
    \vspace{-1em}
\end{figure}

\subsection{AI Tutor Data Pipeline}
A flowchart of our pipeline is shown in Figure \ref{fig:flowchat}. It includes 4 stages: translation, question extraction, RAG, and response generation\footnote{Course materials and source code for the pipeline will be released upon publication.}.

\subsubsection{Translation}
Our pipeline processes two types of inputs: the original English Piazza questions and their human-verified Mandarin translation mentioned in Section 3.1. 
When a question is received in Mandarin, the model translates it into English using the Microsoft Azure AI Translate API\footnote{We chose to build our system using Azure AI Translate API for cost effectiveness. This component is also modular and can be replaced with other APIs.}. We opted for translation over a Mandarin-dedicated LLM to leverage the capabilities of GPT-4, which remains state-of-the-art, especially in programming context.

\subsubsection{Question Extraction (QE)}
A significant portion (about 35\%) of the questions collected from Piazza relate to specific projects, quizzes, and prior exam questions, often referencing only the question number (e.g., "P2 Q10" and "FA22 Exam 2 Q10") in the subject line, without including the question's content in the post. Question extraction (QE) 
stage of the pipeline is equipped to handle such questions. 
Unlike typical computer science projects that require students to implement well-known data structures or algorithms, DS projects involve sequential (numbered) analysis tasks on given datasets, making it important for the model to receive the original question's details. Furthermore, this step is irreplaceable by RAG as both embedding-based and keyword-based retrieval methods face challenges in extracting the exact question from the course materials only using the question number. Embedding-based methods struggle due to the short and numeric nature of these references, while keyword-based systems encounter difficulties due to variability of question number abbreviations (e.g., "Project 2 Question 10" and "Fall 2022 Midterm 2 Question 10").

To address these issues, we used the Llama 3 8B model to generate standardized question numbers from student posts. This model is hosted using Ollama on a machine with a Nvidia GeForce RTX 4090. This link, \url{https://bit.ly/3Lv3Eiw}, provides the few-shot prompt we used for the QE task. The prompt is structured as follows: the task description, the input format accompanied by a brief description of the course materials, the expected output format, and several examples of input and output. The generated standardized question number enables us to accurately fetch the question content from a question bank using a rule-based algorithm. The question bank is created as text files by parsing past exams and solutions (PDFs) and Canvas quizzes using fitz and Canvas API, respectively. 


QE is a crucial step that ensures the subsequent RAG process retrieves the most relevant content related to the specific questions in the student's posts, thereby improving the accuracy and relevancy of the generated responses.

\subsubsection{RAG}
We structured our RAG using \textit{LangChain} \cite{langchain} with a comprehensive vector store of all course materials. This includes lecture notes (Python scripts and Jupyter Notebooks), PowerPoint slides, worksheets (PDFs), complementary readings (Python scripts and Jupyter Notebooks), project and lab repositories (Python scripts, Jupyter Notebooks, and Markdown), and the course website pages (HTML). Various parsers and splitters were employed to manage these diverse data types. For parsing, we converted Jupyter Notebooks to Python scripts using nbconvert, and PowerPoint slides and PDFs to Markdown using pptx2md and fitz. Markdown and HTML content are initially split using corresponding Langchain Markdown and HTML Header Text splitters, and further split by \textit{RecursiveCharacterTextSplitter} in Python for excessively long chunks. Jupyter Notebooks, which are already formatted using Markdown headers, are first split using a modified version of LangChain \textit{MarkdownHeaderTextSplitter} for Jupyter Notebooks and then further split by the \textit{RecursiveCharacterTextSplitter}. We also included the source name in the split chunks to allow reference to course materials in the generated response. The vector store is then created using FAISS \cite{faiss} with OpenAI's text-embedding-ada-002 embedding. 

In similarity search phase, we input both the student post and any extracted questions. We exclude past exams and Canvas quizzes from our search to avoid skewing results towards similar exam or quiz questions, which could prevent the retrieval of lecture materials. We retrieve the top 10 most relevant results.

\subsubsection{Response Generation}

Responses are generated using the Microsoft Azure OpenAI GPT-4 API. This link, \url{https://bit.ly/3SeXvLg}, provides the system and user prompts for the generation task. In the system prompt, we direct the model to function as an AI tutor for non-native English speaker students, stipulating that it should not directly reveal solutions but rather provide hints, explanations, or suggestions. We instruct the model to avoid writing code snippets. We also ask the model to cite RAG results accordingly using the source path. The response is required to be in both English and Mandarin. Any technical vocabulary or error messages that might challenge non-native English speaker students is asked to be compiled into a glossary at the end of the response to aid understanding. In the user prompt, we include the original student post, any extracted question numbers and content, and the results from the RAG.

\section{Results}

\subsection{Examples}


We will present the results from all four stages of our pipeline for one conceptual and one programming question. 

This link, \url{https://bit.ly/3VbHpT8}, illustrates the outcomes for a conceptual post in Mandarin, which asks whether "the Document Object Model (DOM) is a tree or a directed acyclic graph (DAG)". The translation accurately captures the student's query. The question extractor successfully identifies the relevant questions from Spring 22 Exam 2 Question 2 and Quiz 5 Question 2. The RAG retrievals include definitions of trees and DAGs from lecture notes and readings. The response effectively resolves the student's confusion by outlining the definitions shown in the RAG retrievals and clarifying that a tree is a type of DAG.

This link, \url{https://bit.ly/3Ra5Hf2}, displays the results for a programming post asking for tips on P5 Q7. The question extractor accurately identifies the targeted question. The first helpful RAG retrieval provides information about the dataframe, while the second introduces the necessary function from the EDGAR utils module. The response correctly offers general guidance for the question without revealing the solution.

For both questions, the Mandarin translation and technical term glossary were of high quality.


\subsection{Quantitative Results}

\begin{table*}[t]
    \centering
    \begin{tabular}{|c|c|c|c|c|}
         \hline
         & logistics & conceptual & environment setup & programming \\
         \hline
         English & 97.78\% & 90.74\% & 100\% & 97.46\% \\
         \hline
         Mandarin overlap & 98.89\% & 72.22\% & 100\% & 97.97\% \\
         \hline
         Translated overlap & 95.56\% & 74.07\% & 100\% & 97.46\% \\
         \hline 
    \end{tabular}
    \caption{Question extraction accuracy for English posts and of percentages overlaps between QE results for English and Mandarin/translated posts}
    \label{tab:qe}
\end{table*}

\begin{table*}[t]
    \centering
    \begin{tabular}{|c|c|c|c|c|}
         \hline
         & logistics & conceptual & environment setup & programming \\
         \hline
         English relevancy & 6.1 & 7.7 & 7.3 & 6.6 \\
         \hline
         Mandarin overlap & 7.0 & 5.7 & 7.5 & 6.8 \\
         \hline
         Translated overlap & 8.5 & 7.0 & 8.9 & 8.2 \\
         \hline 
    \end{tabular}
    \caption{Number of relevant RAG results for English posts and average numbers of overlaps between RAG results for English and Mandarin/translated posts out of 10 candidates}
    \label{tab:ragrel}
\end{table*}

\begin{table*}[t]
    \centering
    \begin{tabular}{|c|c|c|c|c||c|}
         \hline
         & logistics & conceptual & environment setup & programming & inter-rater reliability \\
         \hline
         Correctness & 80.95\% & 100\% & 100\% & 97.83\% & 94\% \\
         \hline
         Helpfulness & 61.90\% & 93.33\% & 88.89\% & 89.13\% & 85\% \\
         \hline
         Solution presents in response & 0\% & 0\% & 0\% & 13.04\% & 83\% \\
         \hline
         Translation quality & 71.43\% & 86.67\% & 77.78\% & 76.09\% & 67\% \\
         \hline
         Generates technical term glossary & 95.24\% & 86.67\% & 77.78\% & 86.96\% & 87\% \\
         \hline
    \end{tabular}
    \caption{Evaluation scores for generated responses}
    \label{tab:gen}
\end{table*}

All four stages of our pipeline were evaluated by two native Mandarin-speaking authors. In instances of evaluative discrepancies, the third author, who is also a native Mandarin speaker and the course instructor, made the decisive evaluation.

Since we cannot feasibly evaluate all four stages for English, Mandarin, and translated (posts after the translation stage) posts, we focused on evaluating the question extraction (QE) and RAG stages for English posts, which we will demonstrate can serve as the ground truth for evaluating the Mandarin and translated posts. This was achieved by computing the percentage overlap between the English and Mandarin/translated results. We will further demonstrate that the Mandarin posts failed at the RAG stage. Hence, we evaluated the response generation step only with translated posts.

\subsubsection{Translation}
The translation stage showed high quality in translating Mandarin posts to English, answering the first part of RQ1. The Microsoft Azure AI Translation API failed to translate only 2 of the 382 Mandarin posts; these remained untranslated. The translations of all other posts retained their original meanings and all essential information.

\subsubsection{Question Extraction}
The question number extractor can accurately fetch question numbers for all categories of questions, answering the first part of RQ2. The extractor was deemed correct if it outputs the correct question number when the original post (folders, subject, and content) contained sufficient information to identify a specific question. If no specific question could be determined, no question number was output. As indicated in the first row of Table \ref{tab:qe}, the extractor achieved accuracy rates of 97.78\%, 90.74\%, 100\%, and 97.46\% for logistics, conceptual, environment setup, and programming questions, respectively.

The extractor also performs comparably well on Mandarin and translated posts compared to the original English posts. The second and third rows of Table \ref{tab:qe} show the percentages of overlap between the QE results for English and Mandarin/translated posts. The percentages of overlaps between English and Mandarin posts are 98.89\%, 72.22\%, 100\%, and 97.97\% for the four categories, respectively. The percentages of overlaps between English and translated posts are 95.56\%, 74.07\%, 100\%, and 97.46\%. We suspect the high similarity between QE results for the Mandarin and translated posts is due to the fact that question numbers often appear in abbreviations like "P2 Q10", which remain unchanged in both Mandarin and translated posts. Most failed cases occurred in the conceptual category when the original posts did not contained a question number, yet the QE results for the Mandarin and translated posts still attempted to find one.

\subsubsection{RAG}
To assess the RAG and Response Generation stages, we evaluated the top 15\% most viewed question posts and randomly selected 10\% from the remaining posts from each category. This test set include 21 logistics, 14 conceptual, 9 environment setup, and 47 programming questions, totaling 91 posts. This approach ensured coverage of both the questions of highest interest to students and less common queries that might present unique challenges.

We computed two metrics from human evaluation to assess the RAG's performance: relevance and helpfulness. Relevance measured whether the retrieved chunk was pertinent to the question post, while helpfulness determined if the chunk provided an answer or significant contextual information to the question.

RAG cannot replace QE. Firstly, it cannot differentiate between posts that do and do not contain question numbers. Secondly, it does not perform as well as our question number extractor. Among the 91 evaluated posts, 6 logistics, 6 conceptual, and 24 programming posts contain a question number. When RAG searched all course materials including quizzes and exams, it only correctly retrieved 3 (50\%), 4 (66.67\%), and 20 (83.33\%) of the questions within the top 10 searches, respectively. Furthermore, of the 20 correctly retrieved programming questions, only 14 (70\%) are the top 10 RAG results. This confirms that QE is crucial in our pipeline to accommodate the special case of student queries that refer questions by numbers but not content, answering the second part of RQ2. Lastly, including quizzes and exams in RAG searches tends to overwhelm the results for conceptual questions with similar quiz and exam questions, averaging 7.5 out of 10, which obscures the retrieval of relevant lecture materials. Therefore, we exclude quizzes and exams in our RAG search, reserving it only at the QE stage. 


With QE, the RAG results are highly relevant and helpful, which answers RQ3. Among the 10 RAG retrievals, the average number of them being relevant are 6.1, 7.7, 7.3, and 6.6 for logistics, conceptual, environment setup, and programming questions, respectively (first row of Table \ref{tab:ragrel}). Additionally, 80.95\%, 71.43\%, 88.89\%, and 89.36\% of the four categories of question posts contained at least one RAG retrieval that was helpful to students' questions. 

RAG retrievals also performed well with translated posts but less so with Mandarin posts. As shown in the third and fourth rows of Table \ref{tab:ragrel}, out of the 10 retrieved documents for all 382 posts, the average numbers of overlaps between RAG results for English and Mandarin/translated posts are 7.0, 5.7, 7.5, 6.8 and 8.5, 7.0, 8.9, 8.2 for the four categories, respectively. This indicates that translation before the RAG stage helps pick up relevant information, addressing second part of RQ1 by showing that translation is necessary to our AI tutor system where input and output are all in another language.

\subsubsection{Response Generation}
To answer RQ4, we evaluate the generated responses using the following task-specific metrics:

\begin{itemize}[leftmargin=1.5em]
    \item \textbf{Correctness}: Whether the generated output is factually correct and does not contain any misinformation. This metric helps in down voting LLM-generated hallucinations.
    \item \textbf{Helpfulness}: Whether the response effectively addresses the student's question. An answer can be factually correct, but not necessarily situated to the student's question.
    \item \textbf{Solution Presents in Response}: Whether the response directly provides solutions, particularly code snippets, for projects. Our system provides students with hints and directions, not solutions, which we try to achieve by tuning our prompts.
     \item \textbf{Translation Quality}: Whether the translation of the generated output from English to Mandarin is free of errors that could impede understanding. Incomplete translations and missing translations are also considered as errors.
    \item \textbf{Generates Technical Term Glossary}: Whether the generated output includes a glossary of the technical terms related to the question and the response. The prompt sent to the LLM includes instructions for generating such a glossary.
\end{itemize}

Table \ref{tab:gen} presents the evaluation scores for the generated responses, where each metric is binary, scored as 0 or 1. The last column of the table presents the inter-rater reliability for the five generation metrics. These reliability scores measured the initial round of evaluations conducted by two authors and excluding the final adjustments made by the third author in cases of conflict. We utilized Gwet's AC1 to compute the reliability because it is more stable than Cohen's Kappa, particularly when a specific binary rating predominates \cite{wongpakaran2013comparison}. The Gwet's AC1 scores are 94\%, 85\%, 83\%, 67\%, and 87\%, reflecting the consistent evaluation.

In terms of correctness, our pipeline demonstrated a high level of accuracy, achieving perfect scores for both conceptual and environment setup questions (100\% each), and a high score for programming questions (97.83\%). However, the correctness score for logistics questions was lower at 80.95\%. This reduction often occurred in situations where students inquired about specific requirements, such as the necessary formatting for a plot or the use of a particular function. In the absence of clear guidance from the RAG results, the model made assumptions rather than acknowledging the lack of available information.

The helpfulness of the responses, as shown in the second row of Table \ref{tab:gen}, varied slightly across categories but generally remained robust. Responses adequately addressed student questions in 61.90\% of logistics posts, 93.33\% of conceptual posts, 88.89\% of environment setup posts, and 89.13\% of programming posts. 

The responses generally avoided disclosing direct answers. As indicated in the third row, none of the logistics, conceptual, and environment setup questions included direct solutions. However, programming questions experienced a modestly higher rate of direct solution revelation at 13.04\%, primarily because these responses sometimes included parts of code snippets. 


Translation quality is crucial for the comprehension of non-native English speaker students. The pipeline provided Mandarin responses without significant errors in most instances, achieving success rates of 71.43\%, 86.67\%, 77.78\% , and 76.09\% for logistics, conceptual, environment setup, and programming questions, respectively. Failures typically occurred when responses were not fully translated or lacked Mandarin entirely. When translations were complete, they maintained high quality consistently.

The inclusion rate of a technical term glossary is reflected in the final row of Table \ref{tab:gen}. The results demonstrate that 95.24\%, 86.67\%, 77.78\%, and 86.96\% of the queries in the four categories contain a technical term glossary. When the glossary is provided, the terms generated are essential to the questions, uncommonly encountered in everyday English, and their explanations are accurate.


\section{Discussion and Future Work}
\subsection{Our Contributions}
Our multi-source RAG-based AI tutor system, which utilizes diverse course materials and a question extraction mechanism, distinguishes itself from single-sourced models prevalent in prior work. It can generate precise, contextually informed responses, achieving an average helpfulness score of 83.52\% across the evaluated Piazza questions. Additionally, 82.89\% of these responses benefit from the use of question extraction or RAG in generating helpful content.

Over 76.92\% of the responses provide quality responses in Mandarin, and 87.91\% of them contain an effective technical term glossary at the end.
Non-native English speaker (NNES) students may hesitate to approach instructors/TAs due to language challenges and confidence issues. Our system has the potential to mitigate such challenges by offering a non-judgmental, accessible interface, fostering engagement and inclusivity in computing education.

\subsection{Limitations and Future Work}
For future work, we aim to deploy our AI tutor system to NNES students in a real classroom setting after making the following enhancements. We will work on improving accuracy of question extraction, particularly for translated posts, to ensure all target questions are retrieved. Additionally, we will implement safeguards to prevent the system from inadvertently revealing answers, especially in programming questions where unintended code disclosure can occur. To create a more engaging learning experience, the AI tutor will be designed to interact in a conversational manner rather than providing just a single response to posts. After deployment, we plan to assess the questioning behavior and needs of NNES students and tailor our system based on the feedback received.

The release of GPT-4o \cite{gpt4o}, notable for its advancements in vision and non-English language processing, has reduced the cost of vision services. This development enables us to expand our system to accept questions accompanied by images. We will also assess whether our system's performance is enhanced with GPT-4o and consider adapting our system for additional languages to test its validity and extend its applicability.

Given the time-consuming and resource-intensive nature of human evaluation across our system's multiple stages, we will explore automated methods of evaluation to ensure that each component of our pipeline functions efficiently and effectively.

Moreover, we aim to ensure that our AI tutor evolves alongside the course by continuously updating the RAG vector store throughout the semester to incorporate new course materials. Additionally, we will integrate mechanisms to detect when students ask about currently active quiz or exam questions.

Lastly, we intend to develop an interface that allows other instructors to upload their course materials and customize their own AI tutor, further broadening the reach and utility of our pipeline.


\section{Conclusion}
In this work, we developed an AI tutor system that processes student questions in Mandarin and delivers bilingual responses that are correct, helpful, and well-translated, and complete with a technical terms glossary, without directly revealing the answers. Our system utilizes Retrieval-Augmented Generation (RAG) with a special question extraction mechanism to answer queries from a diverse array of course materials, including Jupyter Notebooks, Python scripts, PowerPoint slides, and PDFs. Another main contribution of our AI tutor system is the design of the prompts that instruct the GPT-4 model to apporpriately use resources from question extraction and RAG to provide the appropriate level of guidance in responses. Our evaluation results for our AI tutor on data science questions underscore the potential of AI tutors to enhance educational accessibility for non-native English speaking students in computing courses. Future efforts will focus on refining the system through deployment in real classroom settings and expanding language support to further improve student learning experiences.

\newpage
\bibliographystyle{abbrvnat}
\balance
\bibliography{main}

@String{Computing = "Computing" }

@String{Computer = "{IEEE} Computer" }

@String{Springer = "Springer-Verlag" }

@article{polishchuk2017communication,
  title={Communication barriers faced by English language learners at a university level: factors and solutions},
  author={Polishchuk, Olexander Sergiyovych},
  year={2017},
  publisher={Південноукраїнський національний педагогічний університет імені КД Ушинського}
}

@inproceedings{raj2017students,
  title={What Do Students Feel about Learning Programming Using Both English and Their Native Language?},
  author={Raj, Adalbert Gerald Soosai and Ketsuriyonk, Kasama and Patel, Jignesh M and Halverson, Richard},
  booktitle={2017 International Conference on Learning and Teaching in Computing and Engineering (LaTICE)},
  pages={1--8},
  year={2017},
  organization={IEEE}
}

@inproceedings{NEURIPS2020_6b493230,
 author = {Lewis, Patrick and Perez, Ethan and Piktus, Aleksandra and Petroni, Fabio and Karpukhin, Vladimir and Goyal, Naman and K\"{u}ttler, Heinrich and Lewis, Mike and Yih, Wen-tau and Rockt\"{a}schel, Tim and Riedel, Sebastian and Kiela, Douwe},
 booktitle = {Advances in Neural Information Processing Systems},
 editor = {H. Larochelle and M. Ranzato and R. Hadsell and M.F. Balcan and H. Lin},
 pages = {9459--9474},
 publisher = {Curran Associates, Inc.},
 title = {Retrieval-Augmented Generation for Knowledge-Intensive NLP Tasks},
 url = {https://proceedings.neurips.cc/paper_files/paper/2020/file/6b493230205f780e1bc26945df7481e5-Paper.pdf},
 volume = {33},
 year = {2020}
}

@article{wongpakaran2013comparison,
  title={A comparison of Cohen’s Kappa and Gwet’s AC1 when calculating inter-rater reliability coefficients: a study conducted with personality disorder samples},
  author={Wongpakaran, Nahathai and Wongpakaran, Tinakon and Wedding, Danny and Gwet, Kilem L},
  journal={BMC medical research methodology},
  volume={13},
  pages={1--7},
  year={2013},
  publisher={Springer}
}

@inproceedings{10.1145/3626252.3630909,
author = {Denny, Paul and Leinonen, Juho and Prather, James and Luxton-Reilly, Andrew and Amarouche, Thezyrie and Becker, Brett A. and Reeves, Brent N.},
title = {Prompt Problems: A New Programming Exercise for the Generative AI Era},
year = {2024},
isbn = {9798400704239},
publisher = {Association for Computing Machinery},
address = {New York, NY, USA},
url = {https://doi.org/10.1145/3626252.3630909},
doi = {10.1145/3626252.3630909},
booktitle = {Proceedings of the 55th ACM Technical Symposium on Computer Science Education V. 1},
pages = {296–302},
numpages = {7},
location = {Portland, OR, USA},
series = {SIGCSE 2024}
}

@inproceedings{raj2019impact,
  title={Impact of bilingual cs education on student learning and engagement in a data structures course},
  author={Raj, Adalbert Gerald Soosai and Zhang, Hanqi and Abhyankar, Viren and Mukerjee, Saswati and Zhang, Eda and Williams, Jim and Halverson, Richard and Patel, Jignesh M},
  booktitle={Proceedings of the 19th Koli Calling International Conference on Computing Education Research},
  pages={1--10},
  year={2019}
}

@article{lau2011impact,
  title={The impact of the medium of instruction: The case of teaching and learning of computer programming},
  author={Lau, Wilfred WF and Yuen, Allan HK},
  journal={Education and Information Technologies},
  volume={16},
  pages={183--201},
  year={2011},
  publisher={Springer}
}

@inproceedings{inproceedings,
author = {Guo, Philip},
year = {2018},
month = {04},
pages = {1-14},
title = {Non-Native English Speakers Learning Computer Programming: Barriers, Desires, and Design Opportunities},
doi = {10.1145/3173574.3173970}
}

@misc{opendoors2023,
  title = {Open Doors 2023 Report on International Educational Exchange},
  author = {{Institute of International Education}},
  year = {2023},
  note = {\url{https://opendoorsdata.org/annual-release/}},
}

@inproceedings{10.1145/3626252.3630874,
author = {Shen, Yiyin and Ai, Xinyi and Soosai Raj, Adalbert Gerald and Leo John, Rogers Jeffrey and Syamkumar, Meenakshi},
title = {Implications of ChatGPT for Data Science Education},
year = {2024},
isbn = {9798400704239},
publisher = {Association for Computing Machinery},
address = {New York, NY, USA},
url = {https://doi.org/10.1145/3626252.3630874},
doi = {10.1145/3626252.3630874},
booktitle = {Proceedings of the 55th ACM Technical Symposium on Computer Science Education V. 1},
pages = {1230–1236},
numpages = {7},
location = {Portland, OR, USA},
series = {SIGCSE 2024}
}

@inproceedings{10.1145/3626252.3630789,
author = {Liu, Mengqi and M'Hiri, Faten},
title = {Beyond Traditional Teaching: Large Language Models as Simulated Teaching Assistants in Computer Science},
year = {2024},
isbn = {9798400704239},
publisher = {Association for Computing Machinery},
address = {New York, NY, USA},
url = {https://doi.org/10.1145/3626252.3630789},
doi = {10.1145/3626252.3630789},
booktitle = {Proceedings of the 55th ACM Technical Symposium on Computer Science Education V. 1},
pages = {743–749},
numpages = {7},
location = {Portland, OR, USA},
series = {SIGCSE 2024}
}

@inproceedings{macneil2023experiences,
author = {MacNeil, Stephen and Tran, Andrew and Hellas, Arto and Kim, Joanne and Sarsa, Sami and Denny, Paul and Bernstein, Seth and Leinonen, Juho},
title = {Experiences from Using Code Explanations Generated by Large Language Models in a Web Software Development E-Book},
year = {2023},
isbn = {9781450394314},
publisher = {Association for Computing Machinery},
address = {New York, NY, USA},
url = {https://doi.org/10.1145/3545945.3569785},
doi = {10.1145/3545945.3569785},
booktitle = {Proceedings of the 54th ACM Technical Symposium on Computer Science Education V. 1},
pages = {931–937},
numpages = {7},
location = {Toronto ON, Canada},
series = {SIGCSE 2023}
}

@inproceedings{sarsa2022automatic,
author = {Sarsa, Sami and Denny, Paul and Hellas, Arto and Leinonen, Juho},
title = {Automatic Generation of Programming Exercises and Code Explanations Using Large Language Models},
year = {2022},
isbn = {9781450391948},
publisher = {Association for Computing Machinery},
address = {New York, NY, USA},
url = {https://doi.org/10.1145/3501385.3543957},
doi = {10.1145/3501385.3543957},
booktitle = {Proceedings of the 2022 ACM Conference on International Computing Education Research - Volume 1},
pages = {27–43},
numpages = {17},
location = {Lugano and Virtual Event, Switzerland},
series = {ICER '22}
}

@misc{hicke2023aita,
      title={AI-TA: Towards an Intelligent Question-Answer Teaching Assistant using Open-Source LLMs}, 
      author={Yann Hicke and Anmol Agarwal and Qianou Ma and Paul Denny},
      year={2023},
      eprint={2311.02775},
      archivePrefix={arXiv},
      primaryClass={cs.LG}
}

@inproceedings{10.1145/3626252.3630938,
author = {Liu, Rongxin and Zenke, Carter and Liu, Charlie and Holmes, Andrew and Thornton, Patrick and Malan, David J.},
title = {Teaching CS50 with AI: Leveraging Generative Artificial Intelligence in Computer Science Education},
year = {2024},
isbn = {9798400704239},
publisher = {Association for Computing Machinery},
address = {New York, NY, USA},
url = {https://doi.org/10.1145/3626252.3630938},
doi = {10.1145/3626252.3630938},
booktitle = {Proceedings of the 55th ACM Technical Symposium on Computer Science Education V. 1},
pages = {750–756},
numpages = {7},
location = {Portland, OR, USA},
series = {SIGCSE 2024}
}

@misc{langchain,
  author = {LangChain},
  title = {Applications that can reason. Powered by LangChain.},
  url = {https://www.langchain.com/}
}

@misc{faiss,
  author = {Meta},
  title = {FAISS},
  url = {https://ai.meta.com/tools/faiss/}
}

@misc{gpt4o,
  author = {OpenAI},
  title = {Models},
  url = {https://platform.openai.com/docs/models}
}

@inproceedings{10.1145/3636243.3636252,
author = {Jury, Breanna and Lorusso, Angela and Leinonen, Juho and Denny, Paul and Luxton-Reilly, Andrew},
title = {Evaluating LLM-generated Worked Examples in an Introductory Programming Course},
year = {2024},
isbn = {9798400716195},
publisher = {Association for Computing Machinery},
address = {New York, NY, USA},
url = {https://doi.org/10.1145/3636243.3636252},
doi = {10.1145/3636243.3636252},
booktitle = {Proceedings of the 26th Australasian Computing Education Conference},
pages = {77–86},
numpages = {10},
location = {Sydney, NSW, Australia},
series = {ACE '24}
}

@inproceedings{10.1145/3613904.3642773,
author = {Kazemitabaar, Majeed and Ye, Runlong and Wang, Xiaoning and Henley, Austin Zachary and Denny, Paul and Craig, Michelle and Grossman, Tovi},
title = {CodeAid: Evaluating a Classroom Deployment of an LLM-based Programming Assistant that Balances Student and Educator Needs},
year = {2024},
isbn = {9798400703300},
publisher = {Association for Computing Machinery},
address = {New York, NY, USA},
url = {https://doi.org/10.1145/3613904.3642773},
doi = {10.1145/3613904.3642773},
booktitle = {Proceedings of the CHI Conference on Human Factors in Computing Systems},
articleno = {650},
numpages = {20},
location = {Honolulu, HI, USA},
series = {CHI '24}
}

@article{10.1145/3633287,
author = {Richards, Mike and Waugh, Kevin and Slaymaker, Mark and Petre, Marian and Woodthorpe, John and Gooch, Daniel},
title = {Bob or Bot: Exploring ChatGPT's Answers to University Computer Science Assessment},
year = {2024},
issue_date = {March 2024},
publisher = {Association for Computing Machinery},
address = {New York, NY, USA},
volume = {24},
number = {1},
url = {https://doi.org/10.1145/3633287},
doi = {10.1145/3633287},
journal = {ACM Trans. Comput. Educ.},
month = {jan},
articleno = {5},
numpages = {32}
}

@misc{xie2024gradelikehumanrethinking,
      title={Grade Like a Human: Rethinking Automated Assessment with Large Language Models}, 
      author={Wenjing Xie and Juxin Niu and Chun Jason Xue and Nan Guan},
      year={2024},
      eprint={2405.19694},
      archivePrefix={arXiv},
      primaryClass={cs.AI},
      url={https://arxiv.org/abs/2405.19694}, 
}

\end{document}